\documentclass{article}

\usepackage[preprint,nonatbib]{neurips_2023}

\usepackage[utf8]{inputenc} 
\usepackage[T1]{fontenc}    
\usepackage{hyperref}       
\usepackage{url}            
\usepackage{booktabs}       
\usepackage{amsfonts}       
\usepackage{nicefrac}       
\usepackage{microtype}      
\usepackage{xcolor}         

\usepackage{multirow}
\usepackage{graphicx}
\usepackage{enumitem}
\usepackage{amsmath, amssymb}
\usepackage[font=small]{caption}

\title{MapNeXt: Revisiting Training and Scaling Practices for Online Vectorized HD Map Construction}

\author{%
  Toyota Li \\
  \texttt{toyota.li@foxmail.com} \\
}

\begin{document}

\maketitle

\begin{abstract}
	High-Definition (HD) maps are pivotal to autopilot navigation. Integrating the capability of lightweight HD map construction at runtime into a self-driving system recently emerges as a promising direction. In this surge, vision-only perception stands out, as a camera rig can still perceive the stereo information, let alone its appealing signature of portability and economy. The latest MapTR~\cite{liao2023maptr} architecture solves the online HD map construction task in an end-to-end fashion but its potential is yet to be explored. In this work, we present a full-scale upgrade of MapTR and propose MapNeXt, the next generation of HD map learning architecture, delivering major contributions from the model training and scaling perspectives. After shedding light on the training dynamics of MapTR and exploiting the supervision from map elements thoroughly, MapNeXt-Tiny raises the mAP of MapTR-Tiny from 49.0\% to 54.8\%, without any architectural modifications. Enjoying the fruit of map segmentation pre-training, MapNeXt-Base further lifts the mAP up to 63.9\% that has already outperformed the prior art, a multi-modality MapTR, by 1.4\% while being $\sim1.8\times$ faster. Towards pushing the performance frontier to the next level, we draw two conclusions on practical model scaling: increased query favors a larger decoder network for adequate digestion; a large backbone steadily promotes the final accuracy without bells and whistles. Building upon these two rules of thumb, MapNeXt-Huge achieves state-of-the-art performance on the challenging nuScenes benchmark. Specifically, we push the mapless vision-only single-model performance to be over 78\% for the first time, exceeding the best model from existing methods by 16\%.
\end{abstract}

\section{Introduction}

Autonomous driving is undoubtedly an attractive and challenging field nowadays, where perceiving the environment surrounding the ego-vehicle both accurately and holistically is a crucial pillar. Thus, High-Definition (HD) map with abundant geometric and semantic information is an indispensable ingredient for autopilot. Until recently, offline HD maps constructed with SLAM-based methods~\cite{loam} remain the mainstay of the self-driving community. Though precise and reliable, the global HD maps not only call for massive labor for annotation but also become expensive to maintain once the real-world environment changes. Not surprisingly, it becomes a trend among the industrial enterprises to ``trade perception for HD map''. That is to say,  the ever-increasing focus has been shifted towards building a local HD map on the fly with onboard sensory observations, providing great scalability and timeliness. The rationality of this choice is further backed by the fact that humans could already infer the surrounding scene geometry and semantics straightforwardly based on visual cues, without referring to a map.

Regarding vehicle-mounted sensors, LiDAR is adept at capturing geometry information in the scene, but its return lacks dense details and texture patterns compared to high-resolution cameras. In addition, from the perspective of industry, considering the bulkiness and costliness of LiDAR, mass production of LiDAR-equipped automobiles is hardly accessible. In contrast, the compactness and popularity of cameras render vision-only perception desiderata. Therefore in this work, we prefer to embrace the surrounding-view cameras as the only input source for perception on the vehicular platforms.

Some early attempts formulate online HD map construction as per-pixel prediction of a rasterized map. As a representative work, HDMapNet~\cite{9812383} dissects the entire problem into several sub-tasks, where the principal one is Bird's Eye View (BEV) semantic segmentation while the rest auxiliary tasks supplement instance information by means of complicated post-processing. Virtually, a map element is desired to be defined in a vectorized format, \textit{i.e.}, an \emph{ordered} set of points, that can be readily consumed by subsequent motion forecasting and planning modules. In this spirit, VectorMapNet~\cite{pmlr-v202-liu23ax} evolves the task in hand from dense pixel-level segmentation to sparse instance-level detection, of which the detection head features DETR~\cite{10.1007/978-3-030-58452-8_13} style, characterizing an end-to-end pipeline. However, each vectorized map element is deemed to be in a point sequence form and such an ordered output need to be arranged by an additional auto-regressive model, inevitably dragging down the inference speed. Although modeling online HD map construction as a sparse detection task is a milestone towards simplicity and efficiency, the remaining annoyance is how to elegantly cope with the order of output points. MapTR~\cite{liao2023maptr} stretches this line of research, giving birth to a completely streamlined architecture with parallelized output. Meanwhile, the determination of point order is sidestepped by creating a bunch of permutation-equivalent ground truths to match prediction outputs in diverse orders. The permutations are fulfilled by a collection of geometrical transformations, such as flipping and shifting, which could get rid of the matching ambiguity between many potential network predictions and one ground truth with the specific point order, during both model optimization and evaluation. Consequently, MapTR not only extends the intriguing property of end-to-end execution through to the entire neural network, but also demonstrates performance superiority over precedent models.

We meticulously study the MapTR architecture and suggest that there exist two concerns with respect to its performance. On the one hand, the reasons behind its compelling performance are under-explored. On the other hand, albeit with a breakthrough, the overall performance of MapTR is still unpleasant in real-world applications to date. For these two concerns, we mainly resolve them from the perspective of model \emph{training} and \emph{scaling}:

\emph{Training}. We provide an in-depth analysis of the advanced mechanism of MapTR through the lens of training dynamics and reveal that its improved performance is inherently attributed to augmented ground truths. Unfortunately, this root cause is less publicized by the authors of MapTR. Since the bipartite matching policy from DETR marries a unique prediction among a large pool of (thousands of) queries to one map element, scarce supervision signal is propagated back to the neural network. Augmenting ground truths overcomes the drawbacks of sparse supervision, which is unintentionally materialized in MapTR by equivalent permutations of ground truths. On top of that, we disclose new chances of augmenting the query to permit map elements to be supervised more often, substantially ameliorating the performance. Besides our advance in the map element decoder, we also underscore the vital role of proper domain knowledge transfer during image encoder pre-training. Note that all the above findings are associated with model training schedules, that could plug and play for online HD map construction, without introducing any additional computation cost during inference.

\emph{Scaling}. Foundation models lately make inroads into the vision domain and reigned supreme~\cite{Zhai_2022_CVPR,Liu_2022_CVPR_swin}. For online HD map construction exceptionally, to our best knowledge, we are the first to scale up the model to probe the performance ceiling. Since efficiency is always the first-class citizen in the self-driving kingdom, we also keep in mind that the large-scale architecture design should be still friendly to parallel computing chips. For instance, we shall stick to pure convolution-based image encoders and expand the decoder through the dimension of network width.

To summarize, we present an omni-scale reloading of the MapTR architecture ranging from image encoder to map element decoder based on improved \emph{training} and \emph{scaling} techniques. The proposed architecture is fittingly dubbed as MapNeXt. Our core contributions in this work are threefold:
\begin{enumerate}[leftmargin=1em,topsep=0pt]
	\item Targeted at onboard models, we propose improved training techniques including augmenting the query for map element decoder and preparing dedicated pre-training for image encoder, bringing about striking performance gains without adding any inference budget.
	\item Targeted at offboard models, we offer golden guidelines on model scaling, such as matching the decoder capacity with the quantity of decoding query. We also unveil the feasibility of translating the rapid development of modern image backbones to online HD map learning for the first time.
	\item On the competitive nuScenes benchmark, our real-time MapNeXt-Tiny improves over the strong MapTR baseline by 5\% mAP or so, running even slightly faster; our non-real-time single-modal MapNeXt-Huge sets the new state-of-the-art with a 78.5 mAP, beating the known best multi-modal model by 16 mAP.
\end{enumerate}

We anticipate that our MapNeXt could serve as a promising foundation, prompting more researchers to devote themselves to the arena of online vectorized HD map construction.

\section{Related Work}

\subsection{Vision-only Dynamic Perception for Autopilot}

Moving obstacle perception in autonomous driving is closely related to the techniques of 3D object detection in the field of computer vision. Only with images as the input signal, there mainly exist two branches of research, one is the bottom-up approach, the other is the top-down approach. The former one is represented by CaDDN~\cite{Reading_2021_CVPR} and BEVDet~\cite{2021arXiv211211790H} which maintain the information flow inside the neural network naturally in a feed-forward fashion: extracting 2D image features, explicitly lifting them to the BEV space, and detecting objects in this space. During the process, they mostly employ well-developed existing components (image backbone~\cite{He_2016_CVPR, Lin_2017_CVPR} and 3D object detection head~\cite{Lang_2019_CVPR, Yin_2021_CVPR}), except the specialized PV-to-BEV view transformation module~\cite{10.1007/978-3-030-58568-6_12}. The latter one is represented by DETR3D~\cite{pmlr-v164-wang22b} and PETR~\cite{10.1007/978-3-031-19812-0_31} which cast representative queries as the detected objects and then trace back to the corresponding image features. Specifically, they adopt object-centric queries to aggregate the latent features in the multi-view image space according to the camera intrinsics and extrinsics. Next, these queries are refined in a cascade style with a stack of Transformer layers for the final classification and localization tasks. In general, online HD map construction is resolved similarly to instance detection with BEV representation learning, but the details can be very different.

\subsection{Vision-only Static Perception for Autopilot}

Static element perception in autonomous driving partly refers to semantic map element prediction, a.k.a., online HD map construction. Borrowing the lessons from recent dynamic perception practices, static perception is also established in the BEV space using only vehicle-mounted camera sensors. The online HD Map construction problem is initially solved in a two-step paradigm. For example, the pioneering HDMapNet~\cite{9812383} predicts pixel-wise semantic categories, instance embedding, and direction simultaneously. Heuristic post-processing is necessitated to obtain structural information from the individual dense prediction results. Differently, VectorMapNet~\cite{pmlr-v202-liu23ax} proposes to organize the map elements in a vectorized form that is helpful to downstream tasks~\cite{Gao_2020_CVPR} and regard the problem as set prediction~\cite{10.1007/978-3-030-58452-8_13}, while it still falls into a two-step solution where an auto-regressive generative model is involved. MapTR~\cite{liao2023maptr} first enables end-to-end HD map learning by supervising the hierarchical queries with a variety of permutation-equivalent ground truths. We appreciate the efficiency and impressive performance of MapTR~\cite{liao2023maptr}, so we develop our method on the basis of it.

\subsection{Vision Transformer}

ViT~\cite{dosovitskiy2021an} demonstrates that as a fundamental image encoder architecture, Transformer could deliver on par performance as full-fledged ConvNets~\cite{Liu_2022_CVPR}. DETR~\cite{10.1007/978-3-030-58452-8_13} and several follow-up works~\cite{zhu2021deformable, 10.1007/978-3-031-20077-9_1} demonstrate that as a decoder, with the aid of bipartite matching, Transformer eliminates the necessity of NMS post-processing and thus lends unprecedented succinctness to cutting-edge detectors, without compromising their efficacy. In this work, we migrate DETR-style head to the field of online HD map construction as well, and analyze its behavior mainly through the lens of training dynamics. Part of this work is technically similar to Group DETR~\cite{Chen_2023_ICCV}, but our analysis originates from the introspection of MapTR and we put forward different variants for query augmentation, in complementary to ground truth augmentation.

\section{Approach}

In this section, we first compile a short summary of the task and the most relevant literature. Next, we decompose the model architecture and discuss its components one by one.

\subsection{Preliminary}
\label{sec:preliminary}

Online HD map construction is an emerging topic in the autonomous driving community, which has not been widely investigated. Therefore, we decided to add a little background to provide easy access to a broad audience and ensure that the setup between different papers is consistent. 

Map elements could have dynamic geometrical shape. For example, the lane divider is of open shape while the pedestrian crossing is of closed shape. The nature of varied shape makes it unavailable to model these map elements in a unified parametric manner. Thus, the open-shape and closed-shape map elements are approximated as polylines and polygons respectively, by sampling equidistant points on themselves. Formally, a map element can be discretized into an \emph{ordered} set of points $V=[v_1, v_2, \cdots, v_{N_v}]$, where $N_v$ is the total number of sampled points. This procedure is termed vectorization, so the processed HD map is called vectorized HD map accordingly.

VectorMapNet utilizes an auto-regressive model to sequentially emit the points $v_i, i = 1, 2, \cdots, N_v$ of one map element, which suffers from an inefficient inference. By contrast, MapTR relaxes the constraint of fixed order by expanding each single map element $V$ to a set of permutation-equivalent map elements $\mathcal{V} = (V, \Gamma) = \{V^j=\gamma^j(V), j = 1, 2, \cdots, M\}$, where $\Gamma = \{\gamma^j, j = 1, 2, \cdots, M\}$ denotes a collection of $M$ transformations that reorganize the ordered points of a map element but still reserve both its vectorized format and geometrical shape. In other words, $\forall V^i, V^j \in\mathcal{V}, \mbox{s.t.}, i, j = 1, 2, \cdots, M, i \ne j$, then $V^i$ and $V^j$ are equivalent up to a permutation, described by $\gamma^j \circ (\gamma^i)^{-1}$.

After warming readers up, we shall step into more in-depth understanding and improved design of map element learning in the sequel.

\subsection{Decoder}

\subsubsection{Training}

Following the encoder-decoder paradigm of DETR~\cite{10.1007/978-3-030-58452-8_13}, to frame the map element prediction problem as sparse instance detection, a vast number of object-centric queries $Q = \{q_1, q_2, \cdots, q_N\}$ are created to cover all the map elements in one scene ($N$ is greater than the maximal possible number of map elements).  During inference, MapTR infers all the map elements in one shot and each map element is inferred from an individual query in the decoder. Specifically, each query $q_i \in \mathbb{R}^D, i = 1, 2, \cdots, N$ aggregates spatial information from the encoder feature via cross attention and outputs the corresponding classification and localization predictions $p_i^{(cls)} \in \mathbb{R}^C$ and $p_i^{(loc)} \in \mathbb{R}^2$, where $C$ is the number of pre-defined categories of map elements. For brevity, we combine them into a whole prediction $p_i = [p_i^{(cls)}, p_i^{(loc)}] \in \mathbb{R}^{C+2}$ by concatenation.

Given a set of ground truths $G = \{g_1, g_2, \cdots, g_N\}$ that is padded with $\varnothing$ (no object) to the length of $N$, each ground truth $g_i$ (actually a map element $V_i$ in Section~\ref{sec:preliminary}) is uniquely assigned to one query $q_{\pi(i)} \in Q$ as its label, where $\pi$ is a permutation of the indices. The optimal bipartite matching $\hat{\pi}$ can be determined using the Hungarian algorithm~\cite{RePEc:wly:navlog:v:2:y:1955:i:1-2:p:83-97} by minimizing the overall matching cost between all predictions and all ground truths. Once ground truth labels are defined for each query (as well as its corresponding prediction), the task-specific loss is derived as
\begin{equation}
	\mathcal{L} = \sum_{i=1}^{N} \mathcal{L}_{\rm Hungarian}(p_{\hat{\pi}(i)}, g_i),
\end{equation}
where the Hungarian loss consists of classification, localization, and direction loss as MapTR. For ease of illustration, we only take into account the loss from the final Transformer decoder layer here, while in practice the total loss is summed up from all the decoder layers.

However, considering the limited number of map elements in a self-driving scenario, the bipartite matching strategy makes the neural network receive little ground truth supervision from few learnable queries. In consequence, the convergence speed and detection performance are reduced. Diving deep into MapTR, we notice that the permutation-equivalent transformation mentioned in Section~\ref{sec:preliminary} accidentally addresses this issue to some extent. Concretely speaking, since the queries are believed to be associated with spatial positions~\cite{10.1007/978-3-030-58452-8_13}, by introducing numerous spatially-permuted versions of a map element, the originally one ground truth can be assigned to \emph{distinct} queries now to create more supervised matching pairs. Thus, from our viewpoint, instead of ``stabilizing the learning process'' as claimed by MapTR, the major effect of modeling diverse permutation-equivalent map elements is essentially accelerating the convergence and improving the detection performance, as shown in Figure 5 and Table 2 of the original MapTR paper respectively.

Moreover, in the implementation of MapTR\footnote{\url{https://github.com/hustvl/MapTR}}, \emph{not all the possible} permutations of a map element are performed as stated in the paper. Instead, a \emph{pre-defined} number of $M$ permutations are applied for each map element. In other words, the applied permutations do not cover a full set, but merely a subset with a fixed cardinality $M$. For example, a polygon is only permuted with circular shifting but without reversing its direction in practice. Afterward, random samples are drawn if the total number of permuted samples exceeds the limit $M$. Now, ground truths are augmented as $G = \cup_{j=1}^{M} G^j = \cup_{j=1}^{M} \{g_1^j, g_2^j, \cdots, g_N^j\}$ with $g_i^j = \gamma^j(g_i)$ and the optimal assignment $\hat{\pi}(i, j)$ depends also on the index $j$. The overall loss exactly used by MapTR is calculated as
\begin{equation}
	\mathcal{L} = \sum_{i=1}^{N} \sum_{j=1}^{M} \mathcal{L}_{\rm Hungarian}(p_{\hat{\pi}(i, j)}, g_i^j).
\end{equation}
Observed from the above equations, we further hypothesize that the top performance of MapTR roots in the increment of ground truths, but does not excessively demands a full coverage of \emph{all possible} permuted ground truths. To back up this contradiction, we conduct a controlled experiment using MapTR-Tiny. As shown in Table~\ref{tab:augmented-gt}, the performance difference between a full and partial set of permutations is indeed marginal. In fact, the enhanced supervision furnished by duplicated ground truths is somewhat obscured by the highlighted permutation-invariant ground truth modeling, to which MapTR mostly attributes its success.

\begin{table}[!h]
	\caption{Two different ground truth permutation strategies give rise to similar performance. Each map element is either permuted with all the possibilities (full set) or $M$ randomly sampled transformations (sub set).}
	\label{tab:augmented-gt}
	\centering
	\resizebox{.5\linewidth}{!}{
		\begin{tabular}{lcccc}
			\toprule
			GT permutation & AP\textsubscript{ped} & AP\textsubscript{lane} & AP\textsubscript{road} & mAP \\
			\midrule
			full set (MapTR introduced) & 43.6 & 51.2 & 52.3 & 49.0 \\
			sub set (MapTR implemented) & 44.1 & 51.7 & 51.5 & 49.1 \\
			\bottomrule
		\end{tabular}
	}
\vskip -0.05in
\end{table}

As a corroboration, if MapTR chooses the unordered Chamfer distance cost for point-level matching in the Hungarian algorithm, one query would have the same distance to all the permuted versions of a map element, then the term of localization cost would completely lose its effect of discriminating different permuted ground truths. The merit of augmented localization supervision would be thereby attenuated. Table 8 of the MapTR paper shows that under such a situation, the final performance is lower than permuted ground truths combined with an ordered point-to-point localization cost but still higher than fixed-order ground truths (at least the augmented classification supervision still helps).

Based on the understanding above, it is natural to further augment the supervision for neural networks from the standpoint of query. Mathematically, the original one set of queries is augmented to $K$ sets as $Q = \cup_{k=1}^{K} Q^k = \cup_{k=1}^{K} \{q_1^k, q_2^k, \cdots, q_N^k\}$. Note that during bipartite matching, the ground truths $G$ are still only matched with one set of queries at a time, for the purpose of avoiding post-processing. Therefore, the optimal assignment may be inconsistent across different sets of queries. For the $k$\textsuperscript{th} set of queries, it is represented as $\hat{\pi}_k(i, j)$. Equipped with the augmented queries, the loss is written as
\begin{equation}
	\mathcal{L} = \sum_{k=1}^{K}\sum_{i=1}^{N} \sum_{j=1}^{M} \mathcal{L}_{\rm Hungarian}(p_{\hat{\pi}_k(i, j)}^k, g_i^j).
\end{equation}

Along different axes, augmenting the query could be realized in two modes, sequential or parallel. Resembling the style of DenseNet~\cite{Huang_2017_CVPR}, the sequential mode reuses the queries coming from the previous Transformer decoder layers, so the number of queries will iteratively increase from shallower to deeper Transformer decoder layers. The parallel mode simply uses multiple sets of queries from the very start of Transformer decoder and the same number of queries exists in consecutive Transformer decoder layers.  No matter under which mode, in every Transformer layer, the self-attention interaction only occurs within each individual set of queries. It is noteworthy that the augmented queries are merely used during training and we only apply one set of query $Q^1$ during inference, keeping the deployment latency untouched.

The results are summarized in Table~\ref{tab:augmented-query}. In terms of the sequential mode, iteratively accumulating the last two Transformer layers' queries has exhausted the GPU memory, but only improves the mAP by around 1\%. So we insist on the parallel mode unless otherwise specified. Regarding the parallel mode, we explore the number of additional query set that spans a wide range from 0 to 20. We find that the performance consistently improves with the increasing query set and still does not saturate until running out of GPU memory. The performance advantage of 20 additional sets of query comes at the cost of an enormous memory footprint, impeding our further exploration in the following, so we choose to conservatively enlarge the number of query set to 1+10 by default. Nevertheless, the ever-growing performance in Table~\ref{tab:augmented-query} speaks for the potential of our method to be further unleashed. We expect to witness another performance leap as soon as larger memory is affordable.

\begin{table}
	\centering
	\parbox{.48\linewidth}{
		\centering
		\caption{Augmenting queries under two modes, where the parallel mode is more scalable and effective.}
		\label{tab:augmented-query}
		\resizebox{\linewidth}{!}{
			\begin{tabular}{clcccc}
				\toprule
				mode & \#set of query & AP\textsubscript{ped} & AP\textsubscript{lane} & AP\textsubscript{road} & mAP \\
				\midrule
				sequential & cumulative & 43.0 & 53.2 & 55.0 & 50.4 \\
				\midrule
				\multirow{4}*{parallel} & 1 & 43.6 & 51.2 & 52.3 & 49.0 \\
				& 1+5 & 49.9 & 54.2 & 56.9 & 53.7 \\
				& 1+10 & 51.9 & 54.6 & 55.5 & 54.0 \\
				& 1+20 & 52.7 & 55.8 & 55.9 & 54.8 \\
				\bottomrule
			\end{tabular}
		}
	}\quad
	\parbox{.48\linewidth}{
		\centering
		\caption{Comparison among different options of position embedding. Models are equipped with 1+10 sets of parallel queries as in Table~\ref{tab:augmented-query}, the same in Table~\ref{tab:ffn}.}
		\label{tab:position-embed}
		\resizebox{\linewidth}{!}{
			\begin{tabular}{lcccc}
				\toprule
				position embedding & AP\textsubscript{ped} & AP\textsubscript{lane} & AP\textsubscript{road} & mAP \\
				\midrule
				emb $\rightarrow$ pos & 51.9 & 54.6 & 55.5 & 54.0 \\
				pos $\rightarrow$ emb (sine) & 49.9 & 57.3 & 57.8 & 55.0 \\
				pos $\rightarrow$ emb (linear) & 50.7 & 56.6 & 58.0 & 55.1 \\
				\bottomrule
			\end{tabular}
		}
	}
\vskip -0.2in
\end{table}

In addition, the positional embedding fed to the decoder can be born in different formats. MapTR generates reference locations from implicitly initialized position embedding, which invokes an ambiguity in its geometric meaning, since such an embedding does not have any notion of spatial distribution prior. On the contrary, we advocate producing position embedding with explicitly initialized reference locations. Unlike MapTR that requires a linear projection layer, a sinusoidal encoding function~\cite{NIPS2017_3f5ee243} without trainable parameters is leveraged in our model to map a normalized 2D location into the latent embedding space. As a result, the inference process is even accelerated mildly, from 20.0 to 20.3 FPS on an NVIDIA A100 GPU. We also consider a linear projection layer as an alternative for mapping, but it brings negligible gain. The comparison results are displayed in Table~\ref{tab:position-embed}. Thanks to the positional information straightforwardly injected into the learnable queries, our decoder is optimized in a more easy and interpretable manner. To the best of our knowledge, the most related work might be Anchor DETR~\cite{Wang_Zhang_Yang_Sun_2022}, but our design is simple yet effective in comparison to Anchor DETR: our position encoding is not only exclusive of multiple patterns of anchors, but also leads to a decent performance gain even rejecting neural network layers.

\subsubsection{Scaling}

To scale up, sufficient decoder network capacity is necessary to digest more queries, so as to guarantee an improved performance. For this set of experiments, we use a stronger VoVNetV2 backbone~\cite{Lee_2020_CVPR}, of which the elaboration is deferred to Section~\ref{sec:encoder}. Either trivially increasing the query number or widening the Feed-Forward Network (FFN) independently achieves limited improvement. In stark contrast, combing the two finally yields a high-performing model, as shown in Table~\ref{tab:ffn}. The listed results are non-trivial from the following aspects: \uppercase\expandafter{\romannumeral1}. fixing the FFN dimension to 512, increasing the query brings little gain (line 1\&2), \uppercase\expandafter{\romannumeral2}. after widening the FFN dimension to 1024, the same increment of query ($50\rightarrow75$) brings a larger gain (line 3\&5 vs. line 1\&2), \uppercase\expandafter{\romannumeral3}. fixing the number of query to 50, even widening the FFN dimension to 2048 achieves inferior performance to a balanced combination of 75 instance query and 1024d FFN, albeit with a 3.6M more parameter count (line 4\&5).

\subsection{Encoder}
\label{sec:encoder}

\subsubsection{Training}

MapTR initializes their ResNet~\cite{He_2016_CVPR} or Swin Transformer~\cite{Liu_2021_ICCV_swin} backbone network with ImageNet~\cite{5206848} pre-trained weights, which is an old-fashioned scheme for transfer learning. We perform a pilot study by initializing the ResNet-18 weights pre-trained on CurveLanes~\cite{10.1007/978-3-030-58555-6_41} with CondLaneNet~\cite{Liu_2021_ICCV}, which slightly strengthens its final performance, as exhibited in Table~\ref{tab:pretrain}. Still, one thing worth noting is that \emph{not all kinds of} pre-training helps. The pre-training domain should be as close to our task of interest as possible, \textit{i.e.}, map element learning, in favor of a successful transfer.

To complement this posit, we provide counter examples in Table~\ref{tab:pretrain}. First of all, we adopt a ResNet-50 pre-trained on nuImages with Cascade R-CNN~\cite{Cai_2018_CVPR} as the backbone. This pre-training setting is said to boost up the camera-based BEV 3D object detection significantly~\cite{2022arXiv220405088X}, but it deteriorates the map element construction oppositely, possibly due to the domain gap between the pre-training objectives and the target task. Beyond that, we employ VoVNetV2-99~\cite{Lee_2020_CVPR} as the backbone network that is also pre-trained on ImageNet. We find that a common routine of frozen Batch Normalization (BN) statistics~\cite{pmlr-v37-ioffe15} induces a performance drop of over 1\% mAP. The phenomenon implies that ImageNet might also not be a proper source dataset for our pre-training, since most object-centric images in ImageNet deviate from the driving scenarios. Another potential reason is that ImageNet pre-trained network concentrates on the classification task, then it would take much effort to adapt the weights for other downstream tasks, so it helps less if the target task is more sensitive to localization~\cite{He_2019_ICCV}.

\begin{table}
	\centering
	\parbox{.48\linewidth}{
		\centering
		\caption{Interplay between the number of query and the dimension of FFN. The number of query refers to that of instance query in \emph{a single query set}. All model variants are trained for 110 epochs to full convergence. The specification of timing is in Table~\ref{tab:main}.}
		\label{tab:ffn}
		\resizebox{\linewidth}{!}{
			\begin{tabular}{cccccccc}
				\toprule
				\#query & FFN dim. & \#params. & FPS & AP\textsubscript{ped} & AP\textsubscript{lane} & AP\textsubscript{road} & mAP \\
				\midrule
				50 & 512 & 82.1M & 16.5 & 66.2 & 71.7 & 73.0 & 70.3 \\
				75 & 512 & 82.1M & 16.3 & 66.0 & 72.0 & 73.4 & 70.5 \\
				50 & 1024 & 83.9M & 16.3 & 67.8 & 72.5 & 73.0 & 71.1 \\
				50 & 2048 & 87.6M & 16.3 & 67.1 & 72.4 & 73.5 & 71.0 \\
				75 & 1024 & 84.0M & 16.4 & 68.0 & 73.1 & 74.0 & 71.7 \\
				\bottomrule
			\end{tabular}
		}
	}\quad
	\parbox{.48\linewidth}{
		\centering
		\caption{MapTR-Tiny with a wide spectrum of backbones and pre-training tasks. $^\flat$ indicates neither the shallow network stages nor BN layers are frozen.} 
		\label{tab:pretrain}
		\resizebox{.9\linewidth}{!}{
			\begin{tabular}{cccccc}
				\toprule
				backbone & pretrain & AP\textsubscript{ped} & AP\textsubscript{lane} & AP\textsubscript{road} & mAP \\
				\midrule
				\multirow{2}*{R18} & ImageNet cls$^\flat$ & 39.9 & 49.4 & 48.7 & 46.0 \\
				& CurveLanes & 40.9 & 49.7 & 49.0 & 46.5 \\
				\midrule
				\multirow{2}*{R50} & ImageNet cls$^\flat$  & 44.9 & 52.2 & 53.8 & 50.3 \\
				& nuImages det & 40.4 & 49.3 & 52.0 & 47.3 \\
				\midrule
				\multirow{5}*{V99} & ImageNet cls$^\flat$ & 54.0 & 62.9 & 61.0 & 59.3 \\
				& ImageNet cls$^{\ }$ & 52.0 & 61.1 & 60.5 & 57.9 \\
				& nuScenes det$^{\ }$ & 56.3 & 64.9 & 65.2 & 62.1 \\
				& nuScenes seg$^\flat$ & 58.5 & 65.5 & 67.8 & 63.9  \\
				& nuScenes seg$^{\ }$ & 58.4 & 66.9 & 66.5 & 63.9 \\
				\bottomrule
			\end{tabular}
		}
	}
\vskip -0.2in
\end{table}

All the above analyses taken into account, it is encouraged to pick a highly relevant task and dataset for backbone pre-training, in order to narrow down the domain gap. VoVNetV2-99, initialized with the weights that are successively trained on DDAD-15M with DD3D~\cite{Park_2021_ICCV} and nuScenes with FCOS3D~\cite{Wang_2021_ICCV}, is the \textit{de facto} standard backbone of top-performing 3D object detectors on the nuScenes leaderboard. MapTR armed with the same bespoke weights outstrips its ImageNet pre-trained counterpart by a remarkable 4.2\% mAP. To take one step further, we pre-train the VoVNetV2 backbone on a nuScenes BEV map segmentation task with PETRv2~\cite{Liu_2023_ICCV}, to enjoy the benefit of rich semantic features. As expected, another 1.8\% higher mAP is reached powered by this more relevant pre-training task. By the way, dissimilar to the case in the last paragraph, even frozen BN statistics would not impair the performance of this model, thanks to the same nuScenes data statistics.

\subsubsection{Scaling}

Scaling up the model for online vectorized HD map construction is rarely well-studied. To bridge this gap, we provide a family of MapNeXt variants from on-board to off-board architectures. It should be emphasized that model scaling is never as effortless as expected. For example, a ConvNeXt-XL~\cite{Liu_2022_CVPR} backbone pre-trained on ImageNet performs merely on par with or even inferior to a VoVNetV2-99 pre-trained on DDAD-15M for 3D object detection/map segmentation, despite with a $3.5\times$ parameter amount\footnote{Under the PETRv2 framework for BEV segmentation, ConvNeXt-XL obtains an IoU of 85.6\%/47.6\%/42.5\% for the drive/lane/vehicle class, not as good as the result of 85.6\%/48.9\%/46.4\% achieved by VoVNetV2-99.}. Unlike the above scaling law that plateaus, we reveal that the online HD map construction task fortunately enjoys the profit of large-scale image backbone in Table~\ref{tab:main}, where there still exists no evidence of performance saturation with hundreds of parameters.

\subsection{Neck}

Last but not least, the neck commonly bridges the encoder and decoder part in a detection system, which is instantiated as a PV-to-BEV transformation module in the BEV-oriented detectors. The original publication of MapTR has already sweepingly explored this component, ranging from the classical Inverse Perspective Mapping~(IPM)~\cite{10.1007/BF00201978}, to modern Lift-Splat~\cite{10.1007/978-3-030-58568-6_12}, deformable attention~\cite{10.1007/978-3-031-20077-9_1} and Geometry-guided Kernel Transformer~(GKT)~\cite{2022arXiv220604584C}. We also do not find much difference among these variants during reproduction and simply inherit the GKT module chosen by MapTR.

\section{Main Experiments}

\subsection{Dataset}

nuScenes~\cite{Caesar_2020_CVPR} is a widely-adopted benchmark for versatile autonomous driving tasks. It contains 1, 000 scenes of 20 seconds duration each, of which the key frames are annotated at 2Hz. The entire dataset is split into 700, 150, and 150 scenes for training, validation, and testing respectively. Each sample contains RGB images from 6 surrounding-view cameras, covering a horizontal FOV of $360^\circ$.

Following the convention~\cite{9812383,pmlr-v202-liu23ax,liao2023maptr}, the categories of interest are pedestrian crossing, lane divider and road boundary. Similar to regular object detection tasks, the evaluation metric is also Average Precision (AP). The difference lies in that the distance between two map elements is measured with Chamfer distance (unlike the Intersection over Union between two bounding boxes). For a fair comparison to peer works~\cite{9812383,pmlr-v202-liu23ax,liao2023maptr}, the distance thresholds are ranging from 0.5 to 1.5 with an interval of 0.5.

\subsection{Implementation Details}

In general, we primarily follow MapTR's training protocol. All model architectures are implemented with the PyTorch library~\cite{NEURIPS2019_bdbca288} and the training is distributed on 8 NVIDIA A100 GPU devices with Automated Mixed Precision (AMP)~\cite{micikevicius2018mixed}. The mini-batch size per device is set to 4, except for ConvNeXt-XL and InternImage-H which is halved. The training period lasts for 24 epochs for fast prototyping and 110 epochs for system-level comparison. The initial learning rate is 0.006 and is decayed following a half-cosine-shaped function. The learning rate of the backbone is multiplied by a factor of \nicefrac{1}{10} because it has been pre-trained. We find the final result is fairly robust to the initial learning rate, probably thanks to the AdamW optimizer~\cite{loshchilov2018decoupled}. The weight decay is fixed as 0.01 and the $\ell_2$ norm of gradients is clipped to be no more than 35. The probability of stochastic depth~\cite{10.1007/978-3-319-46493-0_39} applied to ConvNeXt-XL is 0.4. The layer-wise learning rate decay~\cite{bao2022beit} applied to InternImage-H is 0.95. Such regularization is important for the performance of Transformer-like models based on both previous experience~\cite{steiner2022how} and our observations. Data pre-processing is consistent with the precedent practice~\cite{liao2023maptr}, in order to isolate the role of model improvement from other compounding factors. The perception range is [-15.0m, 15.0m] along the X-axis and [-30.0m, 30.0m] along the Y-axis with reference to the ego-vehicle.

\subsection{Quantitative Results}

The model profiling and performance comparison are showcased in Table~\ref{tab:main}. MapNeXt-Tiny outperforms the MapTR-Tiny counterpart by a considerable gain of 5.7\% mAP after training for 24 epochs, with marginally higher throughput. This performance gain holds as a 4.3\% mAP when they are both trained longer until 110 epochs. MapNeXt-Tiny even surpasses MapTR-Base that utilizes Swin Transformer-base~\cite{Liu_2021_ICCV} as the backbone, with only 36\% parameters and 48\% latency. The results prove the effectiveness of our proposed model training strategies. Note that we follow MapTR's training recipe and \emph{do not} tune the hyper-parameters optimized for MapTR, which may even put our MapNeXt at a \emph{disadvantage} in comparison to MapTR.

To verify the generality of MapNeXt, we further replace the image backbone with VoVNetV2-99. The preferable map segmentation pre-training strategy in Table~\ref{tab:pretrain} is applied. The resulting MapNeXt-Base outperforms our own MapNeXt-Tiny by more than 7\% mAP, preserving 81\% of the throughput. When comparing MapNeXt-Base against the best-in-class vision-only MapTR within 24 epochs, it surpasses MapTR-Base in terms of both efficacy and efficiency, \textit{i.e}, 8\% higher mAP and 3.2 more FPS. In addition, we would like to figure out if vision-only MapNeXt-Base could rival the strongest MapTR model, a multi-modality MapTR-Tiny, that takes the best from both worlds of camera and LiDAR. Impressively, the answer is affirmative. MapNeXt-Base outperforms multi-modal MapTR-tiny by an 1.4\% mAP while being 1.8 times fast. The overwhelming performance of MapNeXt-Base justifies the privilege of VoVNet pre-training on nuScenes map segmentation. Moreover, the multi-modal MapTR model harnesses sparse convolution~\cite{s18103337} to process the input LiDAR point cloud, posing challenges to onboard deployment, while our MapNeXt is free of such complex operators.

To scale up, we elevate the model capacity to a magnitude of hundreds of parameters, by upgrading the image encoder to ConvNeXt-XL and InternImage-H. We observe that the corresponding MapNeXt-Large does not level off in performance but widens the gap between itself and multi-modal MapTR to 7.3\% mAP. Echoed with the findings from Table~\ref{tab:ffn}, we increase the query number and FFN dimension together when scaling up the decoder, leading to a non-trivial 2.4\% mAP increment without compromising the inference efficiency. MapNeXt-Large is not a real-time model but can be capitalized on for offboard settings, such as auto-labeling. Importantly, when compared with non-real-time VectorMapNet with only camera input, MapNeXt-Large nearly doubles the accuracy while still running $1.2\times$ faster. It is also worth mentioning that an over-fitting phenomenon has been observed in the later period of training, for which the limited input image resolution ($0.5\times$ resizing) is blamed. Though, to keep a clear comparison of model architecture, we would rather not increase the input scaling factor here but substantiate its effectiveness when participating in the CVPR 2023 challenge afterward in Section~\ref{sec:challenge}. Finally, armed with the InternImage-H~\cite{Wang_2023_CVPR} backbone network, the online HD map construction performance of MapNeXt-Huge is lifted to an unparalleled 78.5\% mAP.

\definecolor{Gray}{gray}{0.5}
\newcommand{\demph}[1]{\textcolor{Gray}{#1}}

\begin{table}
	\caption{State-of-the-art comparison on the nuScenes \texttt{val} set. ``C'' and ``L'' stand for input modality of camera and LiDAR. ``R18/50'', ``V99'' and ``EffNet'' denote ResNet-18/50~\cite{He_2016_CVPR}, VoVNetV2-99~\cite{Lee_2020_CVPR} and EfficientNet~\cite{pmlr-v97-tan19a} respectively. ``PP'' is short for PointPillars~\cite{Lang_2019_CVPR}.  All the latency is measured with a batch size of 1 after warmup. The entries in \demph {gray} are reported by MapTR~\cite{liao2023maptr} while others are timed by ourselves. $^\dagger$ indicates using 75 instance queries and 1024 FFN dimensions, while $^\ddagger$ indicates using 80 instance queries and 2048 FFN dimensions.}
	\label{tab:main}
	\centering
	\resizebox{\linewidth}{!}{
		\begin{tabular}{lccc|cccc|c|cc}
			\toprule
			\multirow{2}*{Architecture}  & \multirow{2}*{Modality} & \multirow{2}*{Backbone}  & \multirow{2}*{Epochs} & \multicolumn{4}{c|}{AP} & \multirow{2}*{\#Param.} &  \multicolumn{2}{c}{FPS} \\
			& & & & ped. & lane & road & avg. & & RTX3090 & A100 \\
			\midrule
			HDMapNet~\cite{9812383} & C & EffNet-B0  & 30 & 14.4 & 21.7 & 33.0 & 23.0 & - & \demph{0.8} & -   \\
			HDMapNet~\cite{9812383} & L & PP & 30 & 10.4 & 24.1 & 37.9 & 24.1 & - & \demph{1.0} & - \\
			HDMapNet~\cite{9812383} & C \& L & EffNet-B0 \& PP & 30 & 16.3 & 29.6 & 46.7 & 31.0 & - & \demph{0.5} & -\\
			\midrule
			VectorMapNet~\cite{pmlr-v202-liu23ax} & C & R50 & 110 & 36.1 & 47.3 & 39.3 & 40.9 & - &  \demph{2.9} & - \\
			VectorMapNet~\cite{pmlr-v202-liu23ax} & L & PP & 110 & 25.7 & 37.6 & 38.6 & 34.0 & - & - & - \\
			VectorMapNet~\cite{pmlr-v202-liu23ax} & C \& L & R50 \& PP & 110 & 37.6 & 50.5 & 47.5 & 45.2 & - & - & - \\
			\midrule
			MapTR-Nano~\cite{liao2023maptr} & C & R18 & 110 & 39.6 & 49.9 & 48.2 & 45.9 & 15.3M & 29.2 & 50.5 \\
			MapTR-Tiny~\cite{liao2023maptr} & C & R50 & 24 & 46.3 & 51.5 & 53.1 & 50.3 & 35.9M & 12.6 & 20.0 \\
			MapTR-Tiny~\cite{liao2023maptr} & C & R50 & 110 & 56.2 & 59.8 & 60.1 & 58.7 & 35.9M & 12.6 & 20.0 \\
			MapTR-Tiny~\cite{liao2023maptr} & C & Swin-Tiny & 24 & 45.2 & 52.7 & 52.3 & 50.1 & 39.9M &  \demph{9.1} & - \\
			MapTR-Small~\cite{liao2023maptr} & C & Swin-Small & 24 & 50.2 & 55.4 & 57.3 & 54.3 & 61.2M &  \demph{7.3} & - \\
			MapTR-Base~\cite{liao2023maptr} & C & Swin-Base & 24 & 50.6 & 58.7 & 58.4 & 55.9 & 99.2M &  \demph{6.1} & - \\
			MapTR-Tiny~\cite{liao2023maptr} & L & SECOND & 24 & 48.5 & 53.7 & 64.7 & 55.6 & - &  \demph{7.2} & - \\
			MapTR-Tiny~\cite{liao2023maptr} & C \& L & R50 \& SECOND & 24 & 55.9 & 62.3 & 69.3 & 62.5 & 39.8M &  5.2 & 6.2 \\
			\midrule
			MapNeXt-Tiny & C & R50 & 24 & 50.3 & 58.8 & 58.7 & 56.0 & 36.0M & 12.7 & 20.3 \\
			MapNeXt-Tiny & C & R50 & 110 & 57.7 & 65.3 & 65.8 & 63.0 & 36.0M & 12.7 & 20.3 \\
			MapNeXt-Base & C & V99 & 24 & 58.5 & 65.5 & 67.8 & 63.9 & 82.1M & 9.3 & 16.5 \\ 
			MapNeXt-Base & C & V99 & 110 & 66.2 & 71.7 & 73.0 & 70.3 & 82.1M & 9.3 & 16.5 \\
			MapNeXt-Base$^\dagger$ & C & V99 & 110 & 67.8 & 73.1 & 74.1 & 71.7 & 84.0M & 9.2 & 16.4 \\
			MapNeXt-Large & C & ConvNeXt-XL & 24 & 66.7 & 72.4 & 70.4 & 69.8 & 360.9M & 3.5 & 5.9 \\
			MapNeXt-Large & C & ConvNeXt-XL & 110 & 71.5 & 74.9 & 74.7 & 73.7 & 360.9M & 3.5 & 5.9 \\
			MapNeXt-Large$^\ddagger$ & C & ConvNeXt-XL & 110 & 73.7 & 78.4 & 76.2 & 76.1 & 366.5M & 3.5 & 5.9 \\
			MapNeXt-Huge$^\ddagger$ & C & InternImage-H & 110 & 77.4 & 79.3 & 78.8 & 78.5 & - & - & - \\
			\bottomrule
		\end{tabular}
	}
\vskip -0.2in
\end{table}

\section{Challenge Results}
\label{sec:challenge}

Built upon the skeleton of MapNeXt, our entry into the online HD map construction competition of \href{https://opendrivelab.com/AD23Challenge.html}{CVPR 2023 End-to-End Autonomous Driving workshop} wins the honorable runner-up with a 73.65 mAP on the test server \emph{without any test time augmentation or model ensemble tricks}, outperforming the official baseline method by a notable 31.4 mAP. Note that the model is merely trained for 24 epochs and our submission is the second earliest one ($\sim$20 days before deadline) on the public leaderboard, so there leaves much room for further enhancement. The dataset of this challenge is set up via reshaping Argoverse2~\cite{NEURIPS_DATASETS_AND_BENCHMARKS2021_4734ba6f}, so the general applicability and potential representation ability of MapNeXt are proved again in a different benchmark.

\section{Conclusion}

In this work, we rethink the optimization hindrance of MapTR and reinforce the training strategies with enriched informative queries. We additionally pinpoint the critical importance of appropriate pre-training. Since the self-driving world prioritizes efficiency, we highlight that these improved training techniques could be inserted into the existing architecture tailored for online HD map construction without interfering in the inference. Besides lightweight onboard models, we also build offboard ones with empirical scaling principles. In summary, we construct the model family named MapNeXt with a wide coverage of onboard and offboard architectures, attaining leading performance while retaining high throughput. Particularly, MapNeXt-Large/Huge sets a new state-of-the-art on the online HD map construction track of the public nuScenes benchmark. We hope this work paves a path to real-world applications of online vectorized HD map construction in autonomous driving.

{\small
	\bibliographystyle{plain}
	\bibliography{egbib}

\begin{thebibliography}{10}

\bibitem{bao2022beit}
Hangbo Bao, Li~Dong, Songhao Piao, and Furu Wei.
\newblock {BE}it: {BERT} pre-training of image transformers.
\newblock In {\em International Conference on Learning Representations}, 2022.

\bibitem{Caesar_2020_CVPR}
Holger Caesar, Varun Bankiti, Alex~H. Lang, Sourabh Vora, Venice~Erin Liong,
  Qiang Xu, Anush Krishnan, Yu~Pan, Giancarlo Baldan, and Oscar Beijbom.
\newblock nuscenes: A multimodal dataset for autonomous driving.
\newblock In {\em Proceedings of the IEEE/CVF Conference on Computer Vision and
  Pattern Recognition (CVPR)}, June 2020.

\bibitem{Cai_2018_CVPR}
Zhaowei Cai and Nuno Vasconcelos.
\newblock Cascade r-cnn: Delving into high quality object detection.
\newblock In {\em Proceedings of the IEEE Conference on Computer Vision and
  Pattern Recognition (CVPR)}, June 2018.

\bibitem{10.1007/978-3-030-58452-8_13}
Nicolas Carion, Francisco Massa, Gabriel Synnaeve, Nicolas Usunier, Alexander
  Kirillov, and Sergey Zagoruyko.
\newblock End-to-end object detection with transformers.
\newblock In Andrea Vedaldi, Horst Bischof, Thomas Brox, and Jan-Michael Frahm,
  editors, {\em Computer Vision -- ECCV 2020}, pages 213--229, Cham, 2020.
  Springer International Publishing.

\bibitem{Chen_2023_ICCV}
Qiang Chen, Xiaokang Chen, Jian Wang, Shan Zhang, Kun Yao, Haocheng Feng, Junyu
  Han, Errui Ding, Gang Zeng, and Jingdong Wang.
\newblock Group detr: Fast detr training with group-wise one-to-many
  assignment.
\newblock In {\em Proceedings of the IEEE/CVF International Conference on
  Computer Vision (ICCV)}, pages 6633--6642, October 2023.

\bibitem{2022arXiv220604584C}
Shaoyu {Chen}, Tianheng {Cheng}, Xinggang {Wang}, Wenming {Meng}, Qian {Zhang},
  and Wenyu {Liu}.
\newblock {Efficient and Robust 2D-to-BEV Representation Learning via
  Geometry-guided Kernel Transformer}.
\newblock {\em arXiv e-prints}, page arXiv:2206.04584, June 2022.

\bibitem{5206848}
Jia Deng, Wei Dong, Richard Socher, Li-Jia Li, Kai Li, and Li~Fei-Fei.
\newblock Imagenet: A large-scale hierarchical image database.
\newblock In {\em 2009 IEEE Conference on Computer Vision and Pattern
  Recognition}, pages 248--255, 2009.

\bibitem{dosovitskiy2021an}
Alexey Dosovitskiy, Lucas Beyer, Alexander Kolesnikov, Dirk Weissenborn,
  Xiaohua Zhai, Thomas Unterthiner, Mostafa Dehghani, Matthias Minderer, Georg
  Heigold, Sylvain Gelly, Jakob Uszkoreit, and Neil Houlsby.
\newblock An image is worth 16x16 words: Transformers for image recognition at
  scale.
\newblock In {\em International Conference on Learning Representations}, 2021.

\bibitem{Gao_2020_CVPR}
Jiyang Gao, Chen Sun, Hang Zhao, Yi~Shen, Dragomir Anguelov, Congcong Li, and
  Cordelia Schmid.
\newblock Vectornet: Encoding hd maps and agent dynamics from vectorized
  representation.
\newblock In {\em Proceedings of the IEEE/CVF Conference on Computer Vision and
  Pattern Recognition (CVPR)}, June 2020.

\bibitem{He_2019_ICCV}
Kaiming He, Ross Girshick, and Piotr Dollar.
\newblock Rethinking imagenet pre-training.
\newblock In {\em Proceedings of the IEEE/CVF International Conference on
  Computer Vision (ICCV)}, October 2019.

\bibitem{He_2016_CVPR}
Kaiming He, Xiangyu Zhang, Shaoqing Ren, and Jian Sun.
\newblock Deep residual learning for image recognition.
\newblock In {\em Proceedings of the IEEE Conference on Computer Vision and
  Pattern Recognition (CVPR)}, June 2016.

\bibitem{Huang_2017_CVPR}
Gao Huang, Zhuang Liu, Laurens van~der Maaten, and Kilian~Q. Weinberger.
\newblock Densely connected convolutional networks.
\newblock In {\em Proceedings of the IEEE Conference on Computer Vision and
  Pattern Recognition (CVPR)}, July 2017.

\bibitem{10.1007/978-3-319-46493-0_39}
Gao Huang, Yu~Sun, Zhuang Liu, Daniel Sedra, and Kilian~Q. Weinberger.
\newblock Deep networks with stochastic depth.
\newblock In Bastian Leibe, Jiri Matas, Nicu Sebe, and Max Welling, editors,
  {\em Computer Vision -- ECCV 2016}, pages 646--661, Cham, 2016. Springer
  International Publishing.

\bibitem{2021arXiv211211790H}
Junjie {Huang}, Guan {Huang}, Zheng {Zhu}, Yun {Ye}, and Dalong {Du}.
\newblock {BEVDet: High-performance Multi-camera 3D Object Detection in
  Bird-Eye-View}.
\newblock {\em arXiv e-prints}, page arXiv:2112.11790, December 2021.

\bibitem{pmlr-v37-ioffe15}
Sergey Ioffe and Christian Szegedy.
\newblock Batch normalization: Accelerating deep network training by reducing
  internal covariate shift.
\newblock In Francis Bach and David Blei, editors, {\em Proceedings of the 32nd
  International Conference on Machine Learning}, volume~37 of {\em Proceedings
  of Machine Learning Research}, pages 448--456, Lille, France, 07--09 Jul
  2015. PMLR.

\bibitem{RePEc:wly:navlog:v:2:y:1955:i:1-2:p:83-97}
H.~W. Kuhn.
\newblock {The Hungarian method for the assignment problem}.
\newblock {\em Naval Research Logistics Quarterly}, 2(1‐2):83--97, March
  1955.

\bibitem{Lang_2019_CVPR}
Alex~H. Lang, Sourabh Vora, Holger Caesar, Lubing Zhou, Jiong Yang, and Oscar
  Beijbom.
\newblock Pointpillars: Fast encoders for object detection from point clouds.
\newblock In {\em Proceedings of the IEEE/CVF Conference on Computer Vision and
  Pattern Recognition (CVPR)}, June 2019.

\bibitem{Lee_2020_CVPR}
Youngwan Lee and Jongyoul Park.
\newblock Centermask: Real-time anchor-free instance segmentation.
\newblock In {\em Proceedings of the IEEE/CVF Conference on Computer Vision and
  Pattern Recognition (CVPR)}, June 2020.

\bibitem{9812383}
Qi~Li, Yue Wang, Yilun Wang, and Hang Zhao.
\newblock Hdmapnet: An online hd map construction and evaluation framework.
\newblock In {\em 2022 International Conference on Robotics and Automation
  (ICRA)}, pages 4628--4634, 2022.

\bibitem{10.1007/978-3-031-20077-9_1}
Zhiqi Li, Wenhai Wang, Hongyang Li, Enze Xie, Chonghao Sima, Tong Lu, Yu~Qiao,
  and Jifeng Dai.
\newblock Bevformer: Learning bird's-eye-view representation from multi-camera
  images via spatiotemporal transformers.
\newblock In Shai Avidan, Gabriel Brostow, Moustapha Ciss{\'e}, Giovanni~Maria
  Farinella, and Tal Hassner, editors, {\em Computer Vision -- ECCV 2022},
  pages 1--18, Cham, 2022. Springer Nature Switzerland.

\bibitem{liao2023maptr}
Bencheng Liao, Shaoyu Chen, Xinggang Wang, Tianheng Cheng, Qian Zhang, Wenyu
  Liu, and Chang Huang.
\newblock Map{TR}: Structured modeling and learning for online vectorized {HD}
  map construction.
\newblock In {\em The Eleventh International Conference on Learning
  Representations}, 2023.

\bibitem{Lin_2017_CVPR}
Tsung-Yi Lin, Piotr Dollar, Ross Girshick, Kaiming He, Bharath Hariharan, and
  Serge Belongie.
\newblock Feature pyramid networks for object detection.
\newblock In {\em Proceedings of the IEEE Conference on Computer Vision and
  Pattern Recognition (CVPR)}, July 2017.

\bibitem{Liu_2021_ICCV}
Lizhe Liu, Xiaohao Chen, Siyu Zhu, and Ping Tan.
\newblock Condlanenet: A top-to-down lane detection framework based on
  conditional convolution.
\newblock In {\em Proceedings of the IEEE/CVF International Conference on
  Computer Vision (ICCV)}, pages 3773--3782, October 2021.

\bibitem{pmlr-v202-liu23ax}
Yicheng Liu, Tianyuan Yuan, Yue Wang, Yilun Wang, and Hang Zhao.
\newblock {V}ector{M}ap{N}et: End-to-end vectorized {HD} map learning.
\newblock In Andreas Krause, Emma Brunskill, Kyunghyun Cho, Barbara Engelhardt,
  Sivan Sabato, and Jonathan Scarlett, editors, {\em Proceedings of the 40th
  International Conference on Machine Learning}, volume 202 of {\em Proceedings
  of Machine Learning Research}, pages 22352--22369. PMLR, 23--29 Jul 2023.

\bibitem{10.1007/978-3-031-19812-0_31}
Yingfei Liu, Tiancai Wang, Xiangyu Zhang, and Jian Sun.
\newblock Petr: Position embedding transformation for multi-view 3d object
  detection.
\newblock In Shai Avidan, Gabriel Brostow, Moustapha Ciss{\'e}, Giovanni~Maria
  Farinella, and Tal Hassner, editors, {\em Computer Vision -- ECCV 2022},
  pages 531--548, Cham, 2022. Springer Nature Switzerland.

\bibitem{Liu_2023_ICCV}
Yingfei Liu, Junjie Yan, Fan Jia, Shuailin Li, Aqi Gao, Tiancai Wang, and
  Xiangyu Zhang.
\newblock Petrv2: A unified framework for 3d perception from multi-camera
  images.
\newblock In {\em Proceedings of the IEEE/CVF International Conference on
  Computer Vision (ICCV)}, pages 3262--3272, October 2023.

\bibitem{Liu_2022_CVPR_swin}
Ze~Liu, Han Hu, Yutong Lin, Zhuliang Yao, Zhenda Xie, Yixuan Wei, Jia Ning, Yue
  Cao, Zheng Zhang, Li~Dong, Furu Wei, and Baining Guo.
\newblock Swin transformer v2: Scaling up capacity and resolution.
\newblock In {\em Proceedings of the IEEE/CVF Conference on Computer Vision and
  Pattern Recognition (CVPR)}, pages 12009--12019, June 2022.

\bibitem{Liu_2021_ICCV_swin}
Ze~Liu, Yutong Lin, Yue Cao, Han Hu, Yixuan Wei, Zheng Zhang, Stephen Lin, and
  Baining Guo.
\newblock Swin transformer: Hierarchical vision transformer using shifted
  windows.
\newblock In {\em Proceedings of the IEEE/CVF International Conference on
  Computer Vision (ICCV)}, pages 10012--10022, October 2021.

\bibitem{Liu_2022_CVPR}
Zhuang Liu, Hanzi Mao, Chao-Yuan Wu, Christoph Feichtenhofer, Trevor Darrell,
  and Saining Xie.
\newblock A convnet for the 2020s.
\newblock In {\em Proceedings of the IEEE/CVF Conference on Computer Vision and
  Pattern Recognition (CVPR)}, pages 11976--11986, June 2022.

\bibitem{loshchilov2018decoupled}
Ilya Loshchilov and Frank Hutter.
\newblock Decoupled weight decay regularization.
\newblock In {\em International Conference on Learning Representations}, 2019.

\bibitem{10.1007/BF00201978}
Hanspeter~A. Mallot, H.~H. B\"{u}lthoff, J.~J. Little, and S.~Bohrer.
\newblock Inverse perspective mapping simplifies optical flow computation and
  obstacle detection.
\newblock {\em Biol. Cybern.}, 64(3):177–185, jan 1991.

\bibitem{micikevicius2018mixed}
Paulius Micikevicius, Sharan Narang, Jonah Alben, Gregory Diamos, Erich Elsen,
  David Garcia, Boris Ginsburg, Michael Houston, Oleksii Kuchaiev, Ganesh
  Venkatesh, and Hao Wu.
\newblock Mixed precision training.
\newblock In {\em International Conference on Learning Representations}, 2018.

\bibitem{Park_2021_ICCV}
Dennis Park, Rares Ambrus, Vitor Guizilini, Jie Li, and Adrien Gaidon.
\newblock Is pseudo-lidar needed for monocular 3d object detection?
\newblock In {\em Proceedings of the IEEE/CVF International Conference on
  Computer Vision (ICCV)}, pages 3142--3152, October 2021.

\bibitem{NEURIPS2019_bdbca288}
Adam Paszke, Sam Gross, Francisco Massa, Adam Lerer, James Bradbury, Gregory
  Chanan, Trevor Killeen, Zeming Lin, Natalia Gimelshein, Luca Antiga, Alban
  Desmaison, Andreas Kopf, Edward Yang, Zachary DeVito, Martin Raison, Alykhan
  Tejani, Sasank Chilamkurthy, Benoit Steiner, Lu~Fang, Junjie Bai, and Soumith
  Chintala.
\newblock Pytorch: An imperative style, high-performance deep learning library.
\newblock In H.~Wallach, H.~Larochelle, A.~Beygelzimer, F.~d\textquotesingle
  Alch\'{e}-Buc, E.~Fox, and R.~Garnett, editors, {\em Advances in Neural
  Information Processing Systems}, volume~32. Curran Associates, Inc., 2019.

\bibitem{10.1007/978-3-030-58568-6_12}
Jonah Philion and Sanja Fidler.
\newblock Lift, splat, shoot: Encoding images from arbitrary camera rigs by
  implicitly unprojecting to 3d.
\newblock In Andrea Vedaldi, Horst Bischof, Thomas Brox, and Jan-Michael Frahm,
  editors, {\em Computer Vision -- ECCV 2020}, pages 194--210, Cham, 2020.
  Springer International Publishing.

\bibitem{Reading_2021_CVPR}
Cody Reading, Ali Harakeh, Julia Chae, and Steven~L. Waslander.
\newblock Categorical depth distribution network for monocular 3d object
  detection.
\newblock In {\em Proceedings of the IEEE/CVF Conference on Computer Vision and
  Pattern Recognition (CVPR)}, pages 8555--8564, June 2021.

\bibitem{steiner2022how}
Andreas~Peter Steiner, Alexander Kolesnikov, Xiaohua Zhai, Ross Wightman, Jakob
  Uszkoreit, and Lucas Beyer.
\newblock How to train your vit? data, augmentation, and regularization in
  vision transformers.
\newblock {\em Transactions on Machine Learning Research}, 2022.

\bibitem{pmlr-v97-tan19a}
Mingxing Tan and Quoc Le.
\newblock {E}fficient{N}et: Rethinking model scaling for convolutional neural
  networks.
\newblock In Kamalika Chaudhuri and Ruslan Salakhutdinov, editors, {\em
  Proceedings of the 36th International Conference on Machine Learning},
  volume~97 of {\em Proceedings of Machine Learning Research}, pages
  6105--6114. PMLR, 09--15 Jun 2019.

\bibitem{NIPS2017_3f5ee243}
Ashish Vaswani, Noam Shazeer, Niki Parmar, Jakob Uszkoreit, Llion Jones,
  Aidan~N Gomez, \L~ukasz Kaiser, and Illia Polosukhin.
\newblock Attention is all you need.
\newblock In I.~Guyon, U.~Von Luxburg, S.~Bengio, H.~Wallach, R.~Fergus,
  S.~Vishwanathan, and R.~Garnett, editors, {\em Advances in Neural Information
  Processing Systems}, volume~30. Curran Associates, Inc., 2017.

\bibitem{Wang_2021_ICCV}
Tai Wang, Xinge Zhu, Jiangmiao Pang, and Dahua Lin.
\newblock Fcos3d: Fully convolutional one-stage monocular 3d object detection.
\newblock In {\em Proceedings of the IEEE/CVF International Conference on
  Computer Vision (ICCV) Workshops}, pages 913--922, October 2021.

\bibitem{Wang_2023_CVPR}
Wenhai Wang, Jifeng Dai, Zhe Chen, Zhenhang Huang, Zhiqi Li, Xizhou Zhu,
  Xiaowei Hu, Tong Lu, Lewei Lu, Hongsheng Li, Xiaogang Wang, and Yu~Qiao.
\newblock Internimage: Exploring large-scale vision foundation models with
  deformable convolutions.
\newblock In {\em Proceedings of the IEEE/CVF Conference on Computer Vision and
  Pattern Recognition (CVPR)}, pages 14408--14419, June 2023.

\bibitem{Wang_Zhang_Yang_Sun_2022}
Yingming Wang, Xiangyu Zhang, Tong Yang, and Jian Sun.
\newblock Anchor detr: Query design for transformer-based detector.
\newblock {\em Proceedings of the AAAI Conference on Artificial Intelligence},
  36(3):2567--2575, Jun. 2022.

\bibitem{pmlr-v164-wang22b}
Yue Wang, Vitor~Campagnolo Guizilini, Tianyuan Zhang, Yilun Wang, Hang Zhao,
  and Justin Solomon.
\newblock Detr3d: 3d object detection from multi-view images via 3d-to-2d
  queries.
\newblock In Aleksandra Faust, David Hsu, and Gerhard Neumann, editors, {\em
  Proceedings of the 5th Conference on Robot Learning}, volume 164 of {\em
  Proceedings of Machine Learning Research}, pages 180--191. PMLR, 08--11 Nov
  2022.

\bibitem{NEURIPS_DATASETS_AND_BENCHMARKS2021_4734ba6f}
Benjamin Wilson, William Qi, Tanmay Agarwal, John Lambert, Jagjeet Singh,
  Siddhesh Khandelwal, Bowen Pan, Ratnesh Kumar, Andrew Hartnett, Jhony
  Kaesemodel~Pontes, Deva Ramanan, Peter Carr, and James Hays.
\newblock Argoverse 2: Next generation datasets for self-driving perception and
  forecasting.
\newblock In J.~Vanschoren and S.~Yeung, editors, {\em Proceedings of the
  Neural Information Processing Systems Track on Datasets and Benchmarks},
  volume~1. Curran, 2021.

\bibitem{2022arXiv220405088X}
Enze {Xie}, Zhiding {Yu}, Daquan {Zhou}, Jonah {Philion}, Anima {Anandkumar},
  Sanja {Fidler}, Ping {Luo}, and Jose~M. {Alvarez}.
\newblock {M$^2$BEV: Multi-Camera Joint 3D Detection and Segmentation with
  Unified Birds-Eye View Representation}.
\newblock {\em arXiv e-prints}, page arXiv:2204.05088, April 2022.

\bibitem{10.1007/978-3-030-58555-6_41}
Hang Xu, Shaoju Wang, Xinyue Cai, Wei Zhang, Xiaodan Liang, and Zhenguo Li.
\newblock Curvelane-nas: Unifying lane-sensitive architecture search and
  adaptive point blending.
\newblock In Andrea Vedaldi, Horst Bischof, Thomas Brox, and Jan-Michael Frahm,
  editors, {\em Computer Vision -- ECCV 2020}, pages 689--704, Cham, 2020.
  Springer International Publishing.

\bibitem{s18103337}
Yan Yan, Yuxing Mao, and Bo~Li.
\newblock Second: Sparsely embedded convolutional detection.
\newblock {\em Sensors}, 18(10), 2018.

\bibitem{Yin_2021_CVPR}
Tianwei Yin, Xingyi Zhou, and Philipp Krahenbuhl.
\newblock Center-based 3d object detection and tracking.
\newblock In {\em Proceedings of the IEEE/CVF Conference on Computer Vision and
  Pattern Recognition (CVPR)}, pages 11784--11793, June 2021.

\bibitem{Zhai_2022_CVPR}
Xiaohua Zhai, Alexander Kolesnikov, Neil Houlsby, and Lucas Beyer.
\newblock Scaling vision transformers.
\newblock In {\em Proceedings of the IEEE/CVF Conference on Computer Vision and
  Pattern Recognition (CVPR)}, pages 12104--12113, June 2022.

\bibitem{loam}
Ji~Zhang and Sanjiv Singh.
\newblock {LOAM:} lidar odometry and mapping in real-time.
\newblock In {\em Robotics: Science and Systems X, University of California},
  2014.

\bibitem{zhu2021deformable}
Xizhou Zhu, Weijie Su, Lewei Lu, Bin Li, Xiaogang Wang, and Jifeng Dai.
\newblock Deformable {DETR}: Deformable transformers for end-to-end object
  detection.
\newblock In {\em International Conference on Learning Representations}, 2021.

\end{thebibliography}
}


\end{document}